\documentclass{article}

    \PassOptionsToPackage{numbers, compress}{natbib}
 \usepackage[dblblindworkshop, final]{neurips_2025}
\workshoptitle{SPACE in Vision, Language, and Embodied AI; and What Makes a Good Video: Next Practices in Video Generation and Evaluation}

\usepackage[utf8]{inputenc} % allow utf-8 input
\usepackage[T1]{fontenc}    % use 8-bit T1 fonts
\usepackage{hyperref}       % hyperlinks
\usepackage{url}            % simple URL typesetting
\usepackage{booktabs}       % professional-quality tables
\usepackage{amsfonts}       % blackboard math symbols
\usepackage{nicefrac}       % compact symbols for 1/2, etc.
\usepackage{microtype}      % microtypography
\usepackage{xcolor}         % colors
\usepackage[pdftex]{graphicx}
\usepackage{subcaption} 
\usepackage{amsmath}
\usepackage{enumitem}
\usepackage{caption}
\title{Seeing Beyond the Scene: Analyzing and Mitigating Background Bias in Action Recognition}

\author{%
  Ellie Zhou\\
  Westmont High School\\
  \texttt{ellie.m.zhou@gmail.com} \\
  \And
  Jihoon Chung \\
  Princeton University \\
  \texttt{jc5933@princeton.edu} \\
  \And
  Olga Russakovsky \\
  Princeton University \\
  \texttt{olgarus@princeton.edu} \\
}
\begin{document}
\maketitle
\begin{abstract}
  Human action recognition models often rely on background cues rather than human movement and pose to make predictions, a behavior known as background bias. We present a systematic analysis of background bias across classification models, contrastive text-image pretrained models, and Video Large Language Models (VLLM) and find that all exhibit a strong tendency to default to background reasoning. Next, we propose mitigation strategies for classification models and show that incorporating segmented human input effectively decreases background bias by 3.78\%. Finally, we explore manual and automated prompt tuning for VLLMs, demonstrating that prompt design can steer predictions towards human-focused reasoning by 9.85\%.
\end{abstract}

\section{Introduction}
Human action recognition aims to identify what the human is doing in the video, but models often rely on the background, rather than human to make predictions, a phenomenon known as background bias ~\cite{choi2019can, chung2022enabling}. E.g., given a video of a person playing violin in a baseball field, the model may predict ``playing baseball'' rather than ``playing violin'' because of the background. This bias arises because datasets often contain consistent correlations between actions and backgrounds (e.g., skiing always occurs on snow), leading models to leverage backgrounds as shortcuts for the action \cite{geirhos2020shortcut}.

Despite the growing popularity of contrastive text-image pre-trained models like CLIP \cite{radford2021learning} and SigLIP2 \cite{tschannen2025siglip} and video large language models (VLLMs) \cite{bai2025qwen25vltechnicalreport,lin2024videollavalearningunitedvisual,zhu2025internvl3}, background bias in these model paradigms has not been extensively studied. Prior research \cite{choi2019can,chung2022enabling,Li_2023_ICCV,wang2021removing} has mainly focused on classification models, models with a fixed vocabulary class prediction. 
% Some works have examined background bias in CLIP and video LLMs. 
Meanwhile, \cite{fukuzawa2025can,wang2024sober, ye2024mm} primarily focus on object classification rather than action recognition, employ simplistic removal-based strategies, and lack a systematic comparison of background bias across model paradigms. While MASH-VLM \cite{bae2025mash} does a great job analyzing scene bias, it is primarily scoped to VLLMs, and they do not quantify how much models rely on human versus background context. Our work fills the gap by performing a comprehensive analysis of background bias across model paradigms while also developing architectural and prompting-based mitigation strategies.

The main contributions are: \\
(1) We analyze background bias across model paradigms -- classification models, contrastive text-image pre-trained models (CLIP, SigLIP2), and VLLMs. We find that all models display background bias, while VLLMs rely less on background cues. \\
(2) We propose strategies to mitigate background bias in classification models, finding that incorporating segmented human inputs reduces background bias. \\
(3) We demonstrate that prompt engineering can effectively steer VLLMs toward human-focused reasoning, with automated prompt tuning emerging as a particularly promising approach.

\begin{table*}[t!]
  \caption{\textbf{Left:} All models contain background bias. \textbf{Right:} CLIP class-level analysis on HAT Action Swap \cite{chung2022enabling}. Bias is highest when backgrounds are distinctive and consistently paired with actions. Swap Human Accuracy (SHAcc) and Swap Background Error (SBErr) are defined in Section 2.}
  \label{table_combined}
  \centering
  \begin{tabular}{c c}
    % ---- Table 1 ----
    \resizebox{0.50\linewidth}{!}{
      \begin{tabular}{llll}
        \toprule
        Model & HAT SHAcc $\uparrow$ & HAT SBErr $\downarrow$ & Mimetics $\uparrow$ \\
        \midrule
        \scriptsize\emph{\textbf{All classes}} \\
        Slow-Only~\cite{feichtenhofer2019slowfast} & \textbf{11.71} & 26.24 & \textbf{6.31}  \\
        CLIP ViT-B/32~\cite{radford2021learning} & 4.32 & \textbf{15.07} & 5.75 \\
        SigLIP2~\cite{tschannen2025siglip} & 4.06 & 20.01 & 4.63 \\
        \midrule
        \scriptsize\emph{\textbf{As 5-choice MCQ}} \\
        Slow-Only~\cite{feichtenhofer2019slowfast} & 35.81 & 55.41 & 57.64 \\
        CLIP ViT-B/32~\cite{radford2021learning} & 29.25 & 53.66 & 46.84 \\
        SigLIP2~\cite{tschannen2025siglip} & 25.46 & 58.91 & 48.95 \\
        *InternVL3-8B~\cite{zhu2025internvl3} & 40.29 & 48.84 & 62.83 \\
        *InternVL3-78B~\cite{zhu2025internvl3} & \textbf{45.73} & \textbf{48.39} & \textbf{66.61} \\
        \bottomrule
      \end{tabular}
    }
    &
    % ---- Table 2 ----
    \resizebox{0.4\linewidth}{!}{
      \begin{tabular}{lrr}
        \toprule
        \multicolumn{3}{c}{High Bias (highest SBErr)} \\
        \midrule
        Background Class & HAT SHAcc $\uparrow$ & HAT SBErr $\downarrow$ \\
        \midrule
        presenting weather forecast   & 3.64 & 89.09 \\
        decorating the christmas tree & 4.62 & 75.38 \\
        cutting pineapple             & 0.00 & 70.37 \\
        ice skating                   & 0.00 & 69.39 \\
        cleaning toilet               & 0.00 & 68.09 \\
        \midrule
        \multicolumn{3}{c}{Low Bias (lowest SBErr)} \\
        \midrule
        Background Class & HAT SHAcc $\uparrow$ & HAT SBErr $\downarrow$ \\
        \midrule
        dancing ballet    & 10.87 & 0.00 \\
        playing clarinet  & 10.91 & 0.00 \\
        washing feet      & 10.91 & 0.00 \\
        waxing chest      & 14.81 & 0.00 \\
        laughing          & 15.52 & 0.00 \\
        \bottomrule
      \end{tabular}
    }
  \end{tabular}
\end{table*}

\vspace{-0.3cm}
\section{Dataset and Metrics}
\vspace{-0.3cm}
To measure background bias, we use HAT Action Swap, introduced in \cite{chung2022enabling}, which contains videos where the human from class A is placed on a mismatched background of class B (e.g., eating ice cream on basketball court). For this benchmark, we report Swap Human Accuracy (SHAcc) and Swap Background Error (SBErr). SHAcc is the fraction of videos where the model correctly predicted the human class (A), and SBErr is the fraction incorrectly predicted as the background class (B). A high SHAcc indicates more reliance on human features, while a high SBErr shows reliance on background. SHAcc and SBErr are directly comparable as they quantify opposite aspects of model behavior, with a decrease in SBErr coinciding with a SHAcc increase and vice versa. We also evaluate on Mimetics \cite{weinzaepfel2021mimetics}, which contains mimed actions without matching scene context, and Kinetics \cite{carreira2017quo}, a human action video dataset. For both Mimetics and Kinetics, we report only standard classification accuracy.
\vspace{-0.4cm}
\section{Analysis of Background Bias}
\vspace{-0.25cm}
\subsection{CLIP and SigLIP models both exhibit strong background bias}
\vspace{-0.25cm}
We analyzed background bias in CLIP ViT-B/32~\cite{radford2021learning} and SigLIP2~\cite{tschannen2025siglip} models, both contrastive text-image pre-trained models known to perform well on visual understanding tasks without task-specific training. We tested on the three ``Random'' Action-Swap mixes from HAT \cite{chung2022enabling} and feed the middle video frame as both models only work on a single image. The 400 action labels from Kinetics were used as text prompts. As Table \ref{table_combined} left shows, both models have a higher SBErr than SHAcc, with CLIP's SHAcc being 4.32\% and SBErr being 15.07\% and SigLIP2 having similar results. This indicates that they are biased towards predicting the background. We broke down results by background class label in Table \ref{table_combined} right. E.g., among all videos which have ``presenting weather forecast'' as background, only 3.64\% predicted the correct human label, while 89.09\% predicted ``presenting weather forecast''. 

High bias tends to occur when background is visually distinctive and consistently paired with the action (e.g., toilet in cleaning toilet). In such cases, the model over-relies on background cues, predicting the background-associated action even when the person’s movements do not match. On the other hand, low bias occurs when background is not distinctive for the action and could appear in many different contexts. E.g., ``playing clarinet'' could take place in a concert hall, outdoors, etc. so there is no distinctive background cue for model to rely on.

\begin{figure}[t]
    \begin{subfigure}[t]{0.35\linewidth}
        \includegraphics[width=\linewidth]{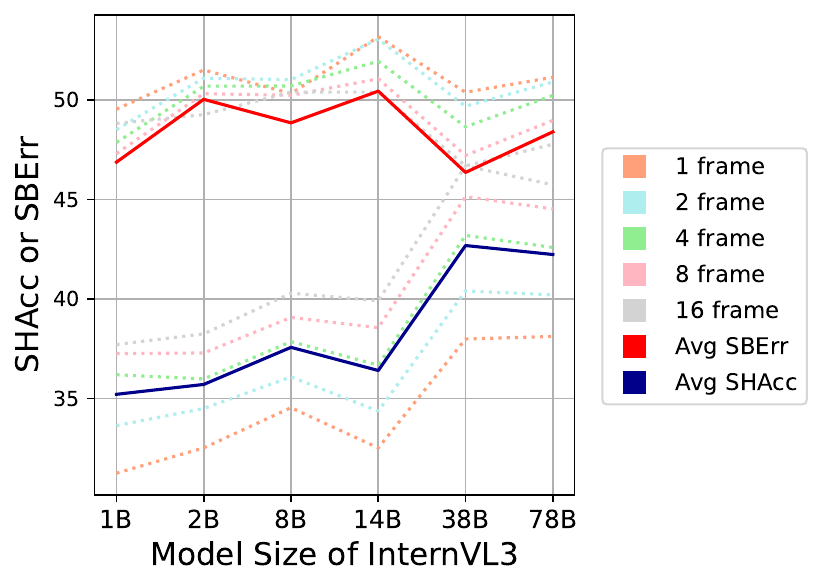}
        \caption{}
        \label{fig:model_size}
    \end{subfigure}
    \begin{subfigure}[t]{0.35\linewidth}
        \includegraphics[width=\linewidth]{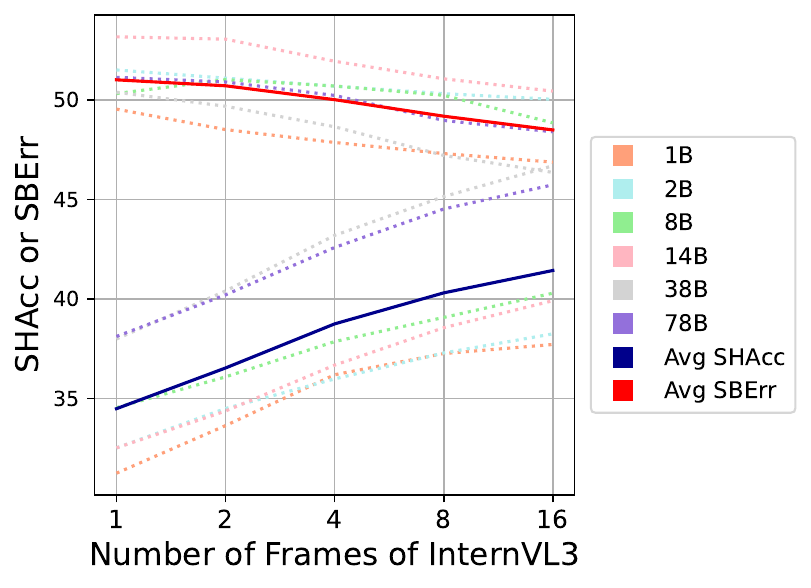}
        \caption{}
        \label{fig:num_frames}
    \end{subfigure}
    \begin{subfigure}[t]{0.27\linewidth}
        \includegraphics[width=\linewidth]{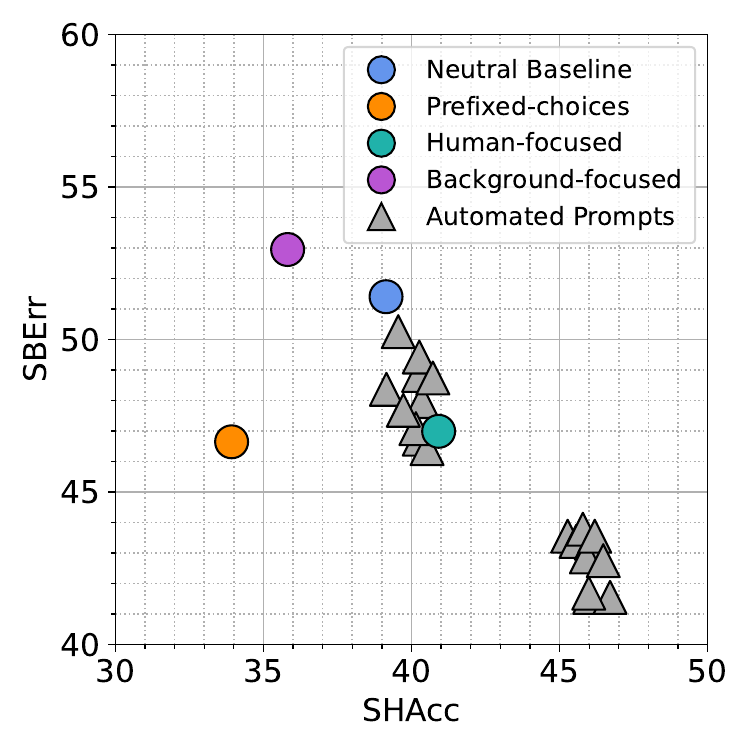}
        \caption{}
        \label{fig:gpt_prompts}
    \end{subfigure}
    \caption{\textbf{(a)} Effect of model size on InternVL3. Increased model capacity improve SHAcc only. \textbf{(b)} Effect of number of frames. Temporal information increases SHAcc and decreases SBErr. \textbf{(c)} Performance of GPT prompts. Automated prompt tuning better reduces background bias.}
    \label{fig:combined}
\end{figure}

\vspace{-0.25cm}
\subsection{Increasing model capacity does not decrease background bias, while temporal input does}
\vspace{-0.25cm}
Seeing that CLIP and SigLIP2 are highly background biased, we asked whether VLLMs, with more parameters and capable of processing multiple frames, could mitigate this bias through their greater capacity or temporal information. To investigate both questions, we varied model capacity and input-frame count independently to isolate each variable’s effect on background bias.

We choose InternVL3 \cite{zhu2025internvl3} for our base model. We prompted InternVL3 with ``What is the action being performed?'' and provided five answer choices: the action label, background label, and three randomly-chosen incorrect Kinetics-400 classes. We use the five-choice format to save tokens and constrain the VLLM to a closed-set classification task, enabling fair comparison with other models. To study performance trends, we varied (1) the number of frames (sampled evenly throughout the video) that we feed to the model and (2) the model size within the InternVL3 family. 

As Figure \ref{fig:model_size} shows, as model size increases, SHAcc improves, but SBErr persists. This suggests that larger model capacity alone is insufficient for robust action understanding. Figure \ref{fig:num_frames} shows that as we give more frames, SHAcc increases, while SBErr decreases. This trend shows that temporal information helps the model focus more on the human motion than the background context. Table \ref{table_combined} left tabulates the results of different models on background bias benchmarks. Overall, VLLMs display background bias but less than both classification model and CLIP/SigLIP2.

\vspace{-0.2cm}
\section{Mitigating Background Bias}
\vspace{-0.4cm}
We explore mitigation solutions in classification models and VLLMs. We construct Kinetics-50 and mini HAT Action Swap, with 50 classes. Section~\ref{section:construction} details how we constructed these.

\begin{table}[t]
    \vspace{-1em}
  \caption{Result for classification model mitigation solutions. Change from Slow-Only baseline is shown in parenthesis. Dual-Branch Sum/Stack and Weighted-Focus show improved accuracy on Kinetics while also mitigating background bias.}
  \label{table_4}
  \centering
  \resizebox{0.6\linewidth}{!}{
      \begin{tabular}{lllll}
        \toprule
        Model & Kinetics-50 $\uparrow$ & HAT SHAcc $\uparrow$ & HAT SBErr $\downarrow$ & Mimetics $\uparrow$ \\
        \midrule
        Slow-Only & 49.93 & 9.62 & 23.42 & 6.87 \\
        Segmented & 23.46\textsubscript{(-26.47)} & 23.34\textsubscript{(+13.72)} & 2.09\textsubscript{(-21.33)} & 9.54\textsubscript{(+2.67)} \\
        Dual-Branch Sum & 52.15\textsubscript{(+2.22)}  & 12.76\textsubscript{(+3.14)} & 20.36\textsubscript{(-3.06)} & 7.85\textsubscript{(+0.98)}\\
        Dual-Branch Stack & 51.51\textsubscript{(+1.58)} & 12.80\textsubscript{(+3.18)} & 19.80\textsubscript{(-3.62)} & 8.28\textsubscript{(+1.41)}\\
        Weighted-Focus & 52.03\textsubscript{(+2.10)} & 12.80\textsubscript{(+3.18)} & 19.64\textsubscript{(-3.78)} & 7.85\textsubscript{(+0.98)} \\
        \bottomrule
      \end{tabular}
    }
\end{table}

\vspace{-0.4cm}
\subsection{Classification models: integrating segmented human input mitigates bias}
\vspace{-0.3cm}

\begin{table}[t!]
\caption{Classification results with and without Places365 augmentation. Augmentation lowers Kinetics-50 accuracy but improves HAT Action-Swap and Mimetics.}
\label{tab:aug_results}
\centering
\resizebox{\textwidth}{!}{%  % scale table to full text width
\begin{tabular}{lllllllll}
    \toprule
    Model & Kinetics-50 $\uparrow$ & Kinetics-50 (+aug) $\uparrow$ & HAT SHAcc $\uparrow$  & HAT SHAcc (+aug) $\uparrow$ & HAT SBErr $\downarrow$ & HAT SBErr (+aug) $\downarrow$ & Mimetics $\uparrow$ & Mimetics (+aug) $\uparrow$ \\
    \midrule
    Slow-Only~\cite{feichtenhofer2019slowfast} & 49.94 & 46.56 & 9.62 & 14.73 & 23.42 & 14.41 & 6.87 & 10.24 \\
    Segmented & 23.46 & 23.26 & 23.34 & 21.89 & 2.09 & 2.17 & 9.54 & 9.82 \\
    Dual-Branch Sum & 52.16 & 39.72 & 12.76 & 24.27 & 20.36 & 7.65 & 7.85 & 9.12 \\
    Dual-Branch Stack & 51.51 & 42.98 & 12.80 & 23.18 & 19.80 & 8.57 & 8.27 & 10.94 \\
    Weighted Focus & 52.03 & 23.86 & 12.76 & 22.54 & 19.64 & 1.85 & 7.85 & 10.10 \\
    \bottomrule
\end{tabular}
} % end resizebox
\end{table}

As Table \ref{table_combined} shows, the Slow-Only classification model \cite{feichtenhofer2019slowfast} relies heavily on background. The most straightforward approach is to eliminate the background entirely. Therefore, as our first mitigation strategy, we applied human segmentation before inputting videos to Slow-Only. As Table \ref{table_4} row 2 shows, feeding human alone leads to improvements in background bias, reflected in HAT SHAcc, HAT SBErr, and Mimetics. However, Kinetics-50 accuracy drops sharply, due to loss of context cues, revealing that background remains important for action understanding. The limitation of the ‘Segmented’ method motivates strategies that retain useful context and mitigate bias. Our remaining methods pursue this balance.

\textbf{Dual-Branch: Sum and Stack} Original video and segmented videos are fed to different Slow-Only, where they are fused in the middle via (1) element-wise addition or (2) channel concatenation.
\textbf{Weighted Focus} We allow the model to adaptively control human vs. background weighting. More details on the model architectures and their training in Section \ref{section:models} and \ref{section:details}.

As shown in Table \ref{table_4}, Dual-Branch Sum/Stack and Weighted Focus improve accuracy on Kinetics-50 while also lowering SBErr, increasing SHAcc, and increasing Mimetics accuracy. This shows that incorporating an additional segmented input effectively improves performance and reduces background bias. These findings highlight an important tradeoff: removing background reduces bias, but hurts accuracy on datasets where action and background are strongly correlated, since useful context is lost. In contrast, retaining background preserves accuracy on such datasets but increases bias. Dual-Branch Sum/Stack and Weighted-Focus achieve an effective balance, improving Kinetics-50 accuracy by up to 2.22\% while reducing background bias by up to 3.78\%.

\vspace{-0.3cm}
\paragraph{Training with Augmented Data} Since background bias stems from consistent scene-action correlations, we designed a targeted augmentation to break these associations. We constructed our own "action-swap" set by segmenting the human from each Kinetics-50 video and combining it with an unrelated background randomly sampled from Places365 \cite{zhou2017places}. The human-background misalignment challenges the model to rely more on human motion and less on background cues. Combining this constructed set with Kinetics-50 doubled the training data size from 24,668 to 49,336 videos. Adding augmented training data led to a trade-off in performance. As Table \ref{tab:aug_results} shows, Kinetics-50 accuracy dropped for all models when trained on augmented data. However, performance on HAT Action Swap and Mimetics improved for all models. This suggests that augmented data improves model robustness to background bias and context-mismatched scenarios - which are emphasized in HAT \cite{chung2022enabling} and Mimetics \cite{weinzaepfel2021mimetics}. Our augmentation de-emphasizes background; since Kinetics’ backgrounds are correlated with the action, removing that cue explains the observed accuracy drop.

\vspace{-0.3cm}
\subsection{Large Language Models: prompt tuning effectively steers predictions}
\vspace{-0.3cm}
Having discovered that VLLMS contain significant background bias, we aimed to reduce their reliance on background. Since architectural modifications are often impractical for large LLMs, we shifted our focus to prompt engineering. We sought to determine if, given LLMs' high sensitivity to text instruction, we could effectively steer the model to focus on human action and reduce background reliance by prompting it in different ways. We used GPT-4o Mini \cite{openai2024gpt4technicalreport}, since it has strong general understanding of language, which would make it well-suited for evaluating different prompts. 

\vspace{-0.3cm}
\paragraph{Hand-Crafted Prompt Tuning}
We first tested hand-crafted prompts with varying levels of guidance. The prompts were as follows (full prompts and performance in Section \ref{section:prompts}): \\
\vspace{-0.5cm}
\begin{itemize}[leftmargin=*, noitemsep, topsep=0pt, parsep=0pt]
    \item \textbf{Neutral baseline} - "\textit{What is the action being performed?}" followed by five action labels.
    \item \textbf{Prefixed-choices} - Same as neutral, but each choice is prefixed by "\textit{a video of a human…}"
    \item \textbf{Human-focused} - Instruct model to consider only the human while ignoring background.
    \item \textbf{Background-focused} -  Instruct model to consider only background while ignoring human.
\end{itemize}

\vspace{-0.1cm}
As shown in Figure \ref{fig:gpt_prompts}, prompting can steer background bias. Specifically, Human-focused prompts improves background bias upon the neutral prompt, while Background-focused worsens the bias. 

\vspace{-0.3cm}
\paragraph{Automated Prompt Tuning}
Manual prompt engineering can reduce background bias to some extent. This raised the question: was this the limit of LLM performance, or simply a limitation of our prompts? To investigate, we turned to automated prompt engineering, which systematically improves prompts through a feedback-driven loop using VLLM. Inspired by \cite{gavrikov2024vision,yang2023large}, we let GPT-4.1 \cite{openai2024gpt4technicalreport} serve as the prompt engineer in a simulated dialogue, repeating the steps below for 20 iterations.

\vspace{-0.1cm}
(1) We first instruct GPT to design a prompt to improve accuracy and reduce background bias. \\
(2) GPT responds with a proposed prompt. \\
(3) We test that prompt on our dataset using GPT 4o-mini and report back the SHAcc and SBErr. \\
(4) GPT uses that feedback to refine its prompt in the next round.

Figure \ref{fig:gpt_prompts} summarizes the performance of the four manually crafted prompts and the 20 automated prompts. Evaluation is done on 75\% of HAT Action Swap, and remaining 25\% is used for automated prompt tuning. Specific prompts and performances are in Section \ref{section:prompts}. While the human-focused manual prompt modestly reduced background bias relative to baseline -- decreasing SBErr by 4.75\%, automated tuning achieved larger gains, decreasing SBErr by up to 9.85\%. These results highlight automated prompt tuning as a more effective bias mitigation strategy than manual prompt design.

\vspace{-0.4cm}
\section{Conclusion}
\vspace{-0.4cm}
In this study, we analyze background bias across classification models, contrastive text-image pre-trained models, and VLLMs and find that the bias is pervasive. We then explore mitigation solutions for classification models and find that integrating both the original and segmented inputs improve Kinetics accuracy and reduce background bias on counterfactual benchmarks like HAT Action-Swap. Finally, for VLLMs, we show that they are sensitive to prompt wording and that background bias can be reduced through carefully designed manual prompts or automated prompt tuning.

\section*{Acknowledgment}

This work was done when Ellie Zhou was a high school intern at Princeton University through Laboratory Learning Program. This work is partially supported by the National Science Foundation under Grant No. 2107048 and 2145198. We thank William Yang for his mentorship throughout the program and Sofia Egan, Alvin Kim, Kathy Lin, and all the Princeton Visual AI lab members for the valuable experiences over the course of this project.  

\medskip

\bibliographystyle{plainnat}
\bibliography{ref}

\newpage
\appendix

\section{Limitations}
While we have shown background bias in both contrastive text-image pre-trained models and video-LLMs, we have only tested on a handful of models and our results may not give the fuller picture of how other models might behave. Similarly, our methods of mitigating background bias were only tested on selected representable models, but we were not able to check if the conclusions hold true for different classification models. Automatic prompt tuning, specifically using LLM as an optimizer, is an actively researched area but still in its experimental stages. For example, while we have seen that some of the prompts were yielding improved results, the improvements were not incremental, where the following prompt is not always better than the previous prompt. While we have mainly used HAT as our background bias benchmark, as the original authors stated, the dataset is synthetically generated and might not follow real-world trends.

\section{Technical Appendices and Supplementary Material}
\subsection{Construction of Kinetics-50 and mini HAT Action Swap}
\label{section:construction}
\textbf{Kinetics-50} To construct the Kinetics 50 dataset, we began with the Kinetics-400 dataset \cite{carreira2017quo} and selected the 50 action classes (same 50 as Mimetics \cite{weinzaepfel2021mimetics}). For the training and validation data, we split the original Kinetics-400 training set (filtered to include only those 50 classes) into 80\% for training and 20\% for validation. For the testing data, we used the original Kinetics-400 validation set, again filtered to include only the 50 selected classes. The final dataset consists of 24,668 training videos, 6167 validation videos, and 2485 test videos. 

\textbf{Mini HAT Action Swap} We use images from the HAT dataset introduced by \cite{chung2022enabling}. They provide pre-generated segmented human videos and inpainted background videos for the Kinetics classes. We used the videos from same 50 classes as above to generate Action-Swap images, where a human figure is combined with a mismatched background drawn from a different action class. This counterfactual setup allowed us to evaluate whether models rely more on human appearance or scene context. The total size of our mini HAT Action-Swap set was 2366 videos.

\subsection{Classification Model Details}
\label{section:models}
\paragraph{Segmented Input} As the most straightforward solution, we remove the background entirely by applying segmentation before feeding the video into Slow-Only \cite{feichtenhofer2019slowfast}. This ensures the model sees only the human and cannot rely on background cues. We first used YOLOv5 \cite{yolov5} to extract object bounding boxes and their confidence scores. From these, the highest-confidence bounding box labeled with the “person” class was selected to be input for SAM2 \cite{ravi2024sam}. We then used SAM2 to propagate the segmentation across all frames to extract consistent human segmentations throughout the video. Non-human regions were set to 0.
\paragraph{Dual-Branch} While segmented input reduces reliance on background cues, it also removes potentially useful context. For example, if the model sees a person swimming, but no water, it may struggle to distinguish between swimming and other similar movements. Therefore, we created a dual-branch architecture that allows the model to learn dynamically from both human and background. The model consists of two parallel streams: one receiving original video and the other receiving segmented video (using same method as previous). Both inputs independently go through the initial layers of Slow-Only (Stem, Stage 1, and Stage 2). After Stage 2, the two feature maps are fused using one of two strategies: (1) Sum: element-wise channel addition, or (2) Stack: concatenation along the channel dimension. The fused representation goes through the remaining layers of the Slow-Only backbone (Stage 3, Stage 4, and Head). In the Stack method, Stage 3 is modified to accept the doubled channel dimension resulting from concatenation. The intuition is that in the early layers, each branch learns low-level features from its input, and after fused, it learns a joint representation integrating both human-focused and contextual cues. 

\paragraph{Weighted Focus} This approach focuses on letting the model adaptively control weighting between human and background. We introduce an auxiliary 3D CNN network that processes the early feature maps of the Slow-Only model. It learns a scalar parameter $\alpha$, which controls relative human-background weighting. To apply this weighting, we use the binary segmentation mask M (where 1 = human, 0 = background) and compute a weighted mask as follows:
\begin{equation}
M_{\text{weighted}} = (1 + \alpha) \cdot M + (1 - \alpha) \cdot (1 - M)
\label{eq:weighted}
\end{equation}
To ensure stability, $\alpha$ is constrained to [-1, 1] using sigmoid, meaning the human region can be scaled up to 2$\times$ or down to 0$\times$, with the background scaled in the opposite direction. The weighted mask is then multiplied with the feature maps, and the modified features continue through the remaining layers of Slow-Only.

\subsection{Classification Model Training Details}
\label{section:details}
All classification models are based on the Slow-only (R50 backbone) variant of the SlowFast model \cite{feichtenhofer2019slowfast}. We sampled 8 evenly spaced frames from each video and applied the same preprocessing transformations as proposed in the original SlowFast paper, including resizing and center-cropping to 224$\times$224.

Models were all trained from scratch on Kinetics-50. Models were trained using the Adam optimizer with a starting learning rate of 0.001. We used a ReduceLROnPlateau scheduler with a patience of 40 epochs and a threshold of 1e-2 to adjust the learning rate dynamically based on validation loss. The batch size was set to 20 for all training runs. All models were trained for 300 epochs.

\subsection{Manual and Automated Prompts for LLM}
\label{section:prompts}
\textbf{Manual Prompts and Performance}
\begin{itemize}
    \item What is the action being performed? \textit{(same prompt for both neutral baseline and prefixed-choices)}\\
    SHAcc: 39.14\%, SBErr: 51.40\% (Neutral Baseline)\\
    SHAcc: 33.93\%, SBErr: 46.65\% (Prefixed-choices)
    \item Focus only on the person and their motion. Ignore the background, scene, or surroundings. Based on the person’s posture, appearance, and movement, what is the action being performed?\\
    SHAcc: 40.92\%, SBErr: 46.99\% 
    \item Please just look at the background and not the person. Based on the background scene, what is the action being performed?\\
    SHAcc: 35.82\%, SBErr: 52.95\% 
\end{itemize}

\textbf{Automated Prompts and Performance}

Below are the 20 generated automated prompts and performance. Best performing prompt is bolded.

\begin{itemize}
    \item Focus only on the person’s movements and actions. What activity is the person doing, regardless of the background?  \\
    SHAcc: 40.24\%, SBErr: 48.83\%
    
    \item Ignore the background. Based only on the person’s movements, what action are they performing?  \\
    SHAcc: 45.27\%, SBErr: 43.55\%
    
    \item Describe only the main action the person is doing, without considering the background or location.  \\
    SHAcc: 40.26\%, SBErr: 49.43\%
    
    \item Based solely on the person’s body movements, what action are they performing in this video? Ignore the background.  \\
    SHAcc: 45.56\%, SBErr: 43.38\%
    
    \item Ignore the setting. What is the person doing, based only on their actions and movements?  \\
    SHAcc: 46.07\%, SBErr: 43.32\%
    
    \item Disregard the background. Identify the action the person is performing by observing their movements only.  \\
    SHAcc: 45.91\%, SBErr: 42.87\%
    
    \item Only consider the person’s actions and body movements. What activity are they doing, without using any clues from the background?  \\
    SHAcc: 39.15\%, SBErr: 48.37\%
    
    \item Focus only on the person’s motion and behavior. What action are they performing, ignoring all background details?  \\
    SHAcc: 46.02\%, SBErr: 41.55\%
    
    \item Watch the person’s movements and actions only. What are they doing, without using any information from the background?  \\
    SHAcc: 40.26\%, SBErr: 46.74\%
    
    \item Based only on the person’s physical actions, what activity are they performing? Do not use any background information.  \\
    SHAcc: 39.55\%, SBErr: 50.26\%
    
    \item Ignore everything except the person’s movements. What action are they performing?  \\
    SHAcc: 45.79\%, SBErr: 43.78\%
    
    \item Looking only at the person’s actions, what are they doing in this video? Ignore the surroundings.  \\
    SHAcc: 40.35\%, SBErr: 47.97\%
    
    \item Ignore the environment. What is the person doing, based only on their actions in the video?  \\
    SHAcc: 46.19\%, SBErr: 43.55\%
    
    \item Focus only on the person’s movements in the video. What action are they performing, without considering the background?  \\
    SHAcc: 40.53\%, SBErr: 46.42\%
    
    \item \textbf{Ignore where the video takes place. What action is the person doing, based only on their movements?  \\
    SHAcc: 46.70\%, SBErr: 41.55\%}
    
    \item Disregard the location and background. What is the person doing, based only on their actions?  \\
    SHAcc: 46.48\%, SBErr: 42.75\%
    
    \item Without using any clues from the background or location, what action is the person performing in this video?  \\
    SHAcc: 40.15\%, SBErr: 47.08\%
    
    \item Ignore the background and setting. What action is the person performing, based only on their movements?  \\
    SHAcc: 45.99\%, SBErr: 41.69\%
    
    \item What is the person doing in this video, based only on their actions and not the background?  \\
    SHAcc: 39.73\%, SBErr: 47.68\%
    
    \item Describe the action the person is performing, using only their movements and ignoring the background.  \\
    SHAcc: 40.72\%, SBErr: 48.74\%
\end{itemize}

\end{document}